\documentclass[conference]{IEEEtran}
\usepackage{times}

\usepackage[numbers]{natbib}
\usepackage{multicol}
\usepackage[bookmarks=true]{hyperref}
\usepackage{graphicx}

\begin{document}

\title{Designing for Fairness in Human-Robot Interactions }


\author{\authorblockN{Houston Claure}
\authorblockA{Department of Computer Science\\Yale University}}


%

\maketitle
    

\IEEEpeerreviewmaketitle

\section{Introduction \& Motivation}
The foundation of successful human collaboration is deeply rooted in the principles of fairness \cite{fehr1999theory}. As robots are increasingly prevalent in various parts of society where they are working alongside groups and teams of humans, their ability to understand and act according to principles of fairness becomes crucial for their effective integration. This is especially critical when robots are part of multi-human teams, where they must make continuous decisions regarding the allocation of resources. These resources can be material, such as tools \cite{gombolay2015decision}, or communicative, such as gaze direction \cite{VazquezTowardsGazeb,mutlu2009footing}, and must be distributed fairly among team members to ensure optimal team performance and healthy group dynamics (see Fig. 1 for an example). Therefore, our research focuses on understanding how robots can effectively participate within human groups by making fair decisions while contributing positively to group dynamics and outcomes.

The growing reliance on intelligent systems emphasizes the importance that fairness plays in their decision-making processes, especially as Artificial Intelligence (AI) technologies are increasingly employed for critical decisions like loan eligibility \cite{bhutta2022much}, university admissions \cite{attaran2018opportunities}, and job interviews \cite{raghavan2020mitigating}. However, it is important to differentiate between the fairness concerns associated with the algorithms referenced above, where their decisions are one-time occurrences, and those made by robots, which require continuous decision-making (e.g., the robot in Fig~1. makes multiple allocation decisions during the collaborative task) \cite{claure2022fairness,claure2021fairness}. Furthermore, unlike the machine learning algorithms that make one-shot decisions, robots are embodied and operate in dynamic environments. This means that robot's actions can lead to harm often not considered in other areas of fairness research, such as physical harm. Examples include self-driving cars that fail to recognize darker-skinned individuals \cite{wilson2019predictive}, caregiving robots that offer unequal support among patients \cite{yew2021trust}, and factory robots that assign tasks unevenly across human workers \cite{ranz2017capability}. Despite these challenges, current research in human-robot interaction (HRI) is largely focused on dyadic interactions between a single robot and a human \cite{chen2018planning}, overlooking the complex dynamics that emerge within larger groups \cite{strohkorb2018ripple, tennent2019micbot}. This gap underscores the need for more research on fairness in robotics, taking into account the continuous and context-dependent nature of robot decision-making in diverse human-robot teams.

My work aims to develop computational
tools and techniques to help overcome the limitations in our understanding of how robots can adopt human concepts of fairness, thereby enhancing their integration as effective team members.. Specifically, my work is focused on three 
research thrusts: 1) building platforms to capture large-scale team
data, 2) expanding our understanding of how unfair robot actions shape
interpersonal relationships, and 3) developing algorithms that embed
fairness into robot decision-making.

\section{Prior Work}

\subsection{Building platforms to capture data from multi-agent groups}
I have devised two novel experimental platforms aimed at exploring the implications of fairness in HRI within the context of group dynamics. These platforms are key assets for the evolution of HRI research. Similar to how simulations have fueled advancements in robotics areas like physical manipulation \cite{nair2022r3m}, easy-to-adapt experimental platforms are essential to facilitate the systematic analysis of social interactions and make strides in our understanding of the role of fairness in HRI.

\begin{figure}
    \centering
    \includegraphics[width=1\linewidth]{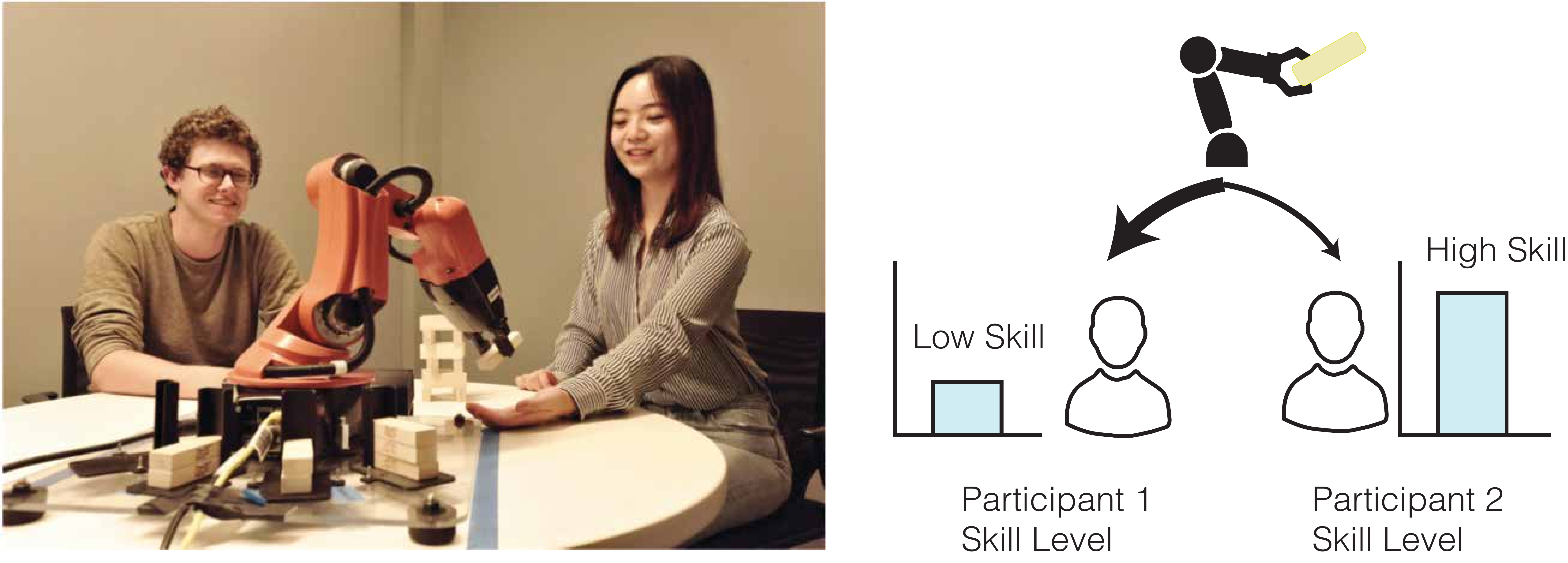}
    \caption{Consider a scenario where a robot has to make decisions on how to distribute resources across two team members. However, one team member is more skilled than the other. How should a robot take skill level into account when dividing its resources to maximize both efficiency and also fairness? This example encapsulates the nuanced responsibility that robots have in human teams: to enhance productivity while fostering an environment of trust and equality.  }
    \label{fig:participants}
    \vspace{-1.5em}
\end{figure}

\subsubsection{Co-Tetris for Online Experiments}  
I developed Co-Tetris as an innovative research tool designed to study collective problem-solving and decision-making. Unlike traditional Tetris \footnote{https://github.com/hbclaure/Co\_Tetris}, which is a single player game, Co-Tetris necessitates cooperation, requiring players to coordinate their actions in real-time. This collaborative twist on the classic game presents a unique way to examine how individuals communicate and strategize towards a shared objective. What makes Co-Tetris a significant contribution is its utility as a research platform. It is a versatile tool that can simulate various collaborative scenarios, offering insights into group dynamics and individual roles within a team. By capturing quantitative data on player performance and behavior, it provides empirical evidence on how people work together under different conditions, including variable difficulties and learning curves. This platform has been instrumental in studies examining fairness and social behavior providing empirical data to inform theories in these domains \cite{claure2020multi,claure2023social}. The adaptability of Co-Tetris to various experimental settings makes it a valuable asset for exploring a wide spectrum of research questions relevant to fairness in human-robot teams.

\subsubsection{Multiplayer Space Invaders for Laboratory Experiments}The Multiplayer Space Invaders Game \cite{claure2024multiplayer} was developed as a platform for analyzing robot decision-making in live, competitive settings. This game extends a traditional Space Invaders game to a multiplayer game where players work to eliminate as many enemy spaceships on their side of the screen as possible. Each player commands an individual spaceship, differentiated by color. The adversaries in the game are represented as alien spaceships and organized into two distinct clusters on the display. A third player can support one of the two players by moving to help eliminate the cluster of enemies on the left or right side. Part of the novelty behind the game's design lies in that it incorporates numerous decision-making moments for the supportive player, providing a framework to analyze different decision patterns and investigate how they evolve temporally. 

\subsection{Building Theory on How Unfair Robot Actions Shape Human Group Dynamics}

In a series of studies -- aimed at exploring the impact of robots on group dynamics and perceptions of fairness \cite{CSCWPaper2024,claure2023social,jung2020robot} -- I investigated how robots' resource distribution behaviors,  a phenomenon I termed ``machine allocation behavior'', affect human interactions. Through the use of a novel tower construction task \cite{jung2020robot} and the Co-Tetris platform \cite{claure2023social}, I analyzed interactions in groups and found that unfair resource allocation by robots could lead to tension among team members and affect team performance, highlighting the importance of fair decisions in human-robot collaboration. A key finding from our experiments suggests that team members who receive more resources from an intelligent machine have feelings of empowerment over their teammates. Expanding on these findings, I sought to predict perceptions of fairness in dynamic interactions. I used the Multiplayer Space Invaders game to demonstrate how fairness judgments evolve over the course of an interaction \cite{CSCWPaper2024} (see Fig.~2). I found that the timing of unfair actions from a robot influences how perceptions of fairness evolve over time. Collectively, these studies underscore the nuanced role of robots in shaping interpersonal dynamics and fairness perceptions, contributing to a deeper understanding of human-robot interaction and informing the design of fair and transparent robotic systems. 

\subsection{Designing Algorithms that Consider Fairness}
This research thrust aims to answer how we can include concepts key to group work, such as fairness, into a robot’s decision-making process. Toward this goal, I explored the problem of a robot distributing resources across team members with varying skill levels through the lens of the multi-armed bandit framework. This led to the development of the strict-rate-constrained upper confidence bound algorithm (UCB) as a viable solution \cite{claure2020multi}. This novel algorithm allows a robot to learn about the skill level of each teammate by observing their performance over time while including a constraint on the minimum rate each human teammate can be selected to receive a resource throughout the course of a task. I provided theoretical guarantees on performance and proved that this algorithm has similar regret bounds to the original unconstrained UCB. To evaluate our algorithm, I deployed the strict rate-constrained UCB on the Co-Tetris platform, adjusting the level of fairness (the number of blocks a lower-skilled player would receive control over).  Our large-scale online study (n = 290 participants) showed that fairness in resource distribution has a significant effect on users’ trust in robots.

\begin{figure}
    \centering
    \includegraphics[width=1\linewidth]{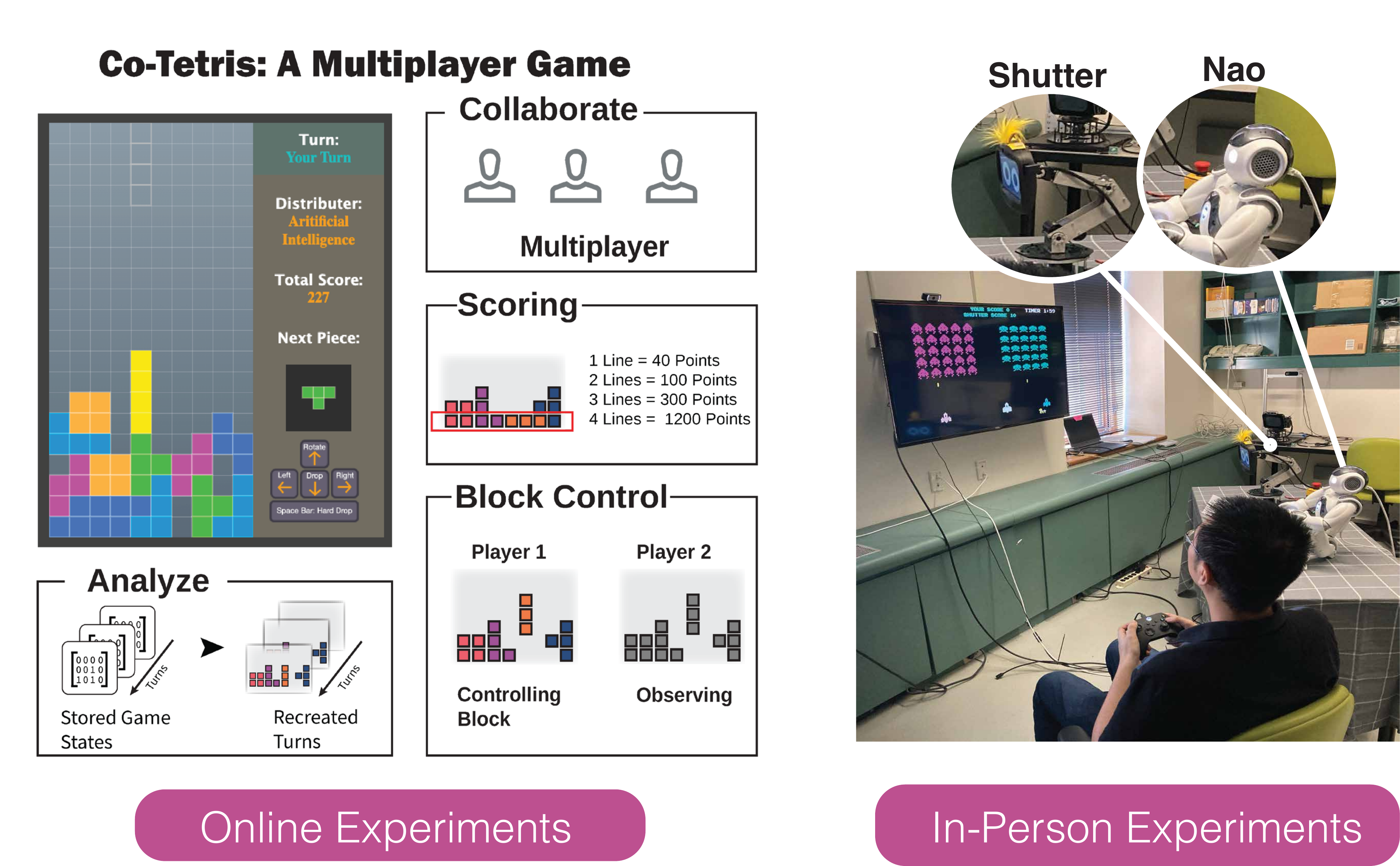}
    \caption{Co-Tetris: a platform for observing group behavior in the face of unfair AI decisions. I ran an experiment where one player was presented as an intelligent system and disproportionately allocated falling Tetris blocks to one player (left). Multiplayer Space Invaders: a robot and a human participant competed to eliminate as many enemies on the screen as possible. A third robot disproportionately supported one player by helping them eliminate enemies for longer periods of time.    }
    \label{fig:participants}
    \vspace{-2em}
\end{figure}

\section{Future Work}
Motivated by people's inherent drive for equity and justice, my research envisions a future where collaborative robots proactively promote fairness within teams, enhancing group harmony and success. This vision encompasses three main objectives: establishing fairness benchmarks in Human-Robot Interaction (HRI) across diverse domains like education, retail, law enforcement, and healthcare; developing methods for learning to enable robots to adapt their actions based on user feedback in real time, thereby aligning closer with human expectations of fairness; and expanding theoretical understanding of fairness in HRI, exploring dynamics of explainability and transparency influenced by cultural and gender factors. Through a multifaceted approach, including online surveys and lab studies, my work aims to develop robots that not only recognize but also rectify disparities, ensuring fair treatment for all team members and fostering effective human-robot collaboration.


\bibliographystyle{plainnat}
\bibliography{paper-template-latex/main}

\end{document}